\documentclass{article}

\usepackage{microtype}
\usepackage{graphicx}
\usepackage{subcaption}
\usepackage{adjustbox}
\usepackage{booktabs} % for professional tables
\usepackage{amsmath}
\usepackage{amsfonts}
\usepackage{dsfont}

\graphicspath{
    {images/}
}

\usepackage{hyperref}

\usepackage[accepted]{icml2019}

\icmltitlerunning{On Physical Adversarial Patches for Object Detection}

\DeclareMathOperator*{\argmax}{arg\,max}

\begin{document}

\twocolumn[
\icmltitle{On Physical Adversarial Patches for Object Detection}

% It is OKAY to include author information, even for blind
% submissions: the style file will automatically remove it for you
% unless you've provided the [accepted] option to the icml2019
% package.

% List of affiliations: The first argument should be a (short)
% identifier you will use later to specify author affiliations
% Academic affiliations should list Department, University, City, Region, Country
% Industry affiliations should list Company, City, Region, Country

% You can specify symbols, otherwise they are numbered in order.
% Ideally, you should not use this facility. Affiliations will be numbered
% in order of appearance and this is the preferred way.
\icmlsetsymbol{equal}{*}

\begin{icmlauthorlist}
\icmlauthor{Mark Lee}{cmu}
\icmlauthor{J. Zico Kolter}{cmu}
\end{icmlauthorlist}

\icmlaffiliation{cmu}{Computer Science Department, Carnegie Mellon University, Pittsburgh, PA 15213, USA}

\icmlcorrespondingauthor{Mark Lee}{marklee@andrew.cmu.edu}

% You may provide any keywords that you
% find helpful for describing your paper; these are used to populate
% the "keywords" metadata in the PDF but will not be shown in the document
\icmlkeywords{Machine Learning, ICML}

\vskip 0.3in
]

% this must go after the closing bracket ] following \twocolumn[ ...

% This command actually creates the footnote in the first column
% listing the affiliations and the copyright notice.
% The command takes one argument, which is text to display at the start of the footnote.
% The \icmlEqualContribution command is standard text for equal contribution.
% Remove it (just {}) if you do not need this facility.

\printAffiliationsAndNotice{}  % leave blank if no need to mention equal contribution
% \printAffiliationsAndNotice{\icmlEqualContribution} % otherwise use the standard text.

\begin{abstract}
In this paper, we demonstrate a physical adversarial patch attack against object detectors, notably the YOLOv3 detector.  Unlike previous work on physical object detection attacks, which required the patch to overlap with the objects being misclassified or avoiding detection, we show that a properly designed patch can suppress virtually \emph{all} the detected objects in the image.  That is, we can place the patch anywhere in the image, causing all existing objects in the image to be missed entirely by the detector, even those far away from the patch itself.  This in turn opens up new lines of physical attacks against object detection systems, which require no modification of the objects in a scene.  A demo of the system can be found at \small{\url{https://youtu.be/WXnQjbZ1e7Y}}.

% We introduce a simple adversarial patch attack for suppressing YOLOv3 detections. Prior patch attacks either required patch pixels to be unbounded (thus physically infeasible) or required the patch to overlap the target objects. Our method synthesizes bounded-pixel patches capable of suppressing detections in the whole image when placed at random locations and orientations, and also outperforms prior methods in the unbounded case. We also demonstrate a physical attack by printing our patch and fooling YOLOv3 running real-time on webcam input. 
\end{abstract}

\section{Introduction}
\label{introduction}

This paper considers the creation of adversarial patches against object detection systems. Broadly, adversarial patch attacks refer to a class of attacks on machine learning systems that add some ``patch'' or perturbation to the image causing the system to mislabel the image. Unlike traditional adversarial examples, they are not imperceptible, but modify the image in a way that should \emph{not} change the underlying output according to human intuition. Past work has demonstrated the feasibility of these attacks (including in physical settings) in the context of classification \cite{adversarial-patch,kurakin2018adversarial} and object detection \cite{eykholt2018robust,xie2017adversarial,surveillance}. However, in the object detection setting, these attacks have required a user to manipulate the object being attacked itself, i.e., by placing the patch over the object.

We present an alternative (and we believe, stronger) adversarial patch attack against object detection. Specifically, we construct a physical adversarial patch that, when placed in a image, suppresses \emph{all} objects previously detected in the image, even those that are relatively far away from the patch. The techniques we use to design the patch are relatively straightforward applications of existing techniques: projected gradient descent approaches \cite{kurakin2016adversarial,madry2017towards} followed by expectation over transformations \cite{athalye2018synthesizing}, specifically optimizing a loss that we believe to be well-suited to object detection systems.

We demonstrate our attack on the YOLOv3 architecture, robustly suppressing detections over a wide range of positions for the object. We illustrate the power of the method both on the COCO dataset, where we evaluate the mAP of the system after the attack, and with a physical attack against YOLOv3 running in real-time on webcam input. The possibility of such attacks opens up new threat vectors for many machine learning systems. For example, it suggests it would be possible to suppress the detection of \emph{all} objects for an autonomous car's vision system (e.g. pedestrians, other cars, street signs), not by requiring us to manipulate each object, but just by placing a well-crafted sign on the sidewalk.

\section{Related Work}

The field of adversarial attacks against machine learning systems is broad enough at this point that we focus here only on the related work most closely related to our approach.

\subsection{Adversarial Patch for Classification}
\label{sec:adversarial-patch}
Adversarial patch attacks were first introduced by \cite{adversarial-patch} for image classifiers. The goal is to produce localized, robust, and universal perturbations that are applied to an image by masking instead of adding pixels. The patch found by \cite{adversarial-patch} is able to fool multiple ImageNet models into predicting ``toaster" whenever the patch is in view, even in physical space as a printed sticker.  However, because classification systems only classify each image as a single class, to some extent this attack relies on the fact that it can simply place a high-confidence ``deep net toaster'' into an image (even if it does not look like a toaster to humans) and override other classes in the image.

\subsection{Adversarial Patches for Image Segmentation}

Because of the limitations of the classification setting, several other works have investigated the use of adversarial patches in the object detection setting \cite{sharif2016accessorize,eykholt2018robust,adversarial-glasses,surveillance,shapeshifter,adversarial-face}. However, for the few cases in this domain dealing with physical adversarial examples, virtually all focused on the creation of an object that \emph{overlaps} the object of interest, to either change its class or suppress detection. In contrast, our approach looks specifically at adversarial patches that do not overlap the objects of interest in the scene.

The work that bears the most similarity to our own is the DPatch method \cite{dpatch}, which explicitly creates patches that do not overlap with the objects of interest. However, the DPatch method was only tested on digital images, and contains a substantial flaw that makes it unsuitable for real experiments: the patches produced in the DPatch work are never clipped to the allowable image range (i.e., clipping colors to the $[0,1]$ range) and thus do not correspond to actual perturbed images. Furthermore, it is not trivial to use the DPatch loss to obtain valid adversarial images: we compare this approach to our own and show that we are able to generate substantially stronger attacks.

% \subsection{DPatch}
% \label{sec:dpatch}
% DPatch extends adversarial patch to object detection. For the ``targeted" attack, the goal is to fool the detector into recognizing only $( \delta_x, \delta_y, \delta_w, \delta_h, l)$ as the bounding box, representing the $x$ ,$y$, width, height, and target label of the patch. The ``untargeted" case simply sets the target label $l$ to be $0$. 

% The paper only demonstrates an ``unclipped" attack, i.e. pixels can be outside of the $[0,1]$ range. This means patches found by this method are not adversarial in physical space, as pixels outside of $[0,1]$ cannot be printed, and as unclipped patches do not generally remain adversarial after clipping.

\subsection{YOLO}
YOLO is a ``one-shot" object detector with state-of-the-art performance on certain metrics running up to $3 \times$ faster than other models \cite{yolov3}. 
It treats the input image as an $S \times S$ grid, each cell predicting $B$ bounding boxes and their confidence scores; and each box predicting $C$ class probabilities, conditioned on there being an object in the box.  We specifically use the YOLOv3 model as the object detection system we use for our demonstrations, though other object detectors would be possible as well. 

\section{Methodology}
\label{sec:methodology}

\subsection{Notation}
\label{sec:notation}
Let $h_{\theta}$ denote a hypothesis function with parameters $\theta$ defining the model (layers, weights, etc); $x$ denote some input to $h_{\theta}$ with a corresponding target of $y$; and $J(h_{\theta}(x), y)$ denote a loss function mapping predictions made by the hypothesis $h_{\theta}$ on input $x$ and the target $y$ to some real-valued number.

\subsection{Attack Formulation}

Here we present our methodology for creating adversarial patches for object detection.  Note that the methods here are based upon existing work: specifically untargeted PGD with expection over transformation, but the results suggest that these attacks are surprisingly stronger than previously thought. We consider the following mathematical formulation of finding an adversarial patch:
$$
\argmax_{\delta} \mathbb{E}_{(x,y) \sim \mathcal{D}, t \sim \mathcal{T}} [J(h_{\theta}(A(\delta, x, t)), y)] \label{eq:optimization}
$$
where $\mathcal{D}$ is a distribution over samples, $T$ is a distribution over patch transformations (to be discussed shortly), and $A$ is a ``patch application function'' that transforms the patch $\delta$ with $t$ and applies the result to the image $x$ by masking the appropriate pixels.  Note that the maximization over $\delta$ is done outside the expectation, i.e., we are considering a class of ``universal'' adversarial perturbations.

The DPatch method attempts to solve a similar objective by minimizing the loss for a carefully crafted target $\hat y$ as described in \cite{dpatch}, performing the update:
\begin{equation}
\delta := \delta - \alpha \nabla_{\delta} J(h_{\theta}(A(\delta, x, t)), \hat y) \label{eq:dpatch-method}
\end{equation}
While this update works fairly well for fitting patches in the digital space, our experiments show that patches found in this way are weakly adversarial when a box-constraint is applied, requiring many update iterations and consistently plateauing at a relatively high mAP (see \autoref{fig:clipped_untargeted_plots}).
Reasons we believe DPatch fails are elaborated in \autoref{sec:why-it-works}. 

Instead, we adopt a simpler approach and simply take the optimization problem at face value and maximize the loss for the original targets $y$ directly for samples and transformations drawn from $\mathcal{D}$ and $T$ respectively.  This is essentially just the standard untargeted PGD approach \cite{madry2017towards}, originally introduced as the Basic Iterative Method \cite{kurakin2016adversarial}, with expectation over transformation \cite{athalye2018synthesizing} applied to the patch itself.
The update does not push the patch $\delta$ towards any particular target label or bounding box. This contrasts with the DPatch update in \autoref{eq:dpatch-method} which requires a target label in $\hat y$ for both the untargeted and targeted cases; this is generally a non-issue as our goal is to suppress detections.  Also following past work, we consider a normalized steepest ascent method under the $\ell_\infty$ norm, which results in the update
\begin{equation}
\delta := \text{clip}_{[0,1]}(\delta + \alpha \cdot \text{sign}(\nabla_{\delta} J(h_{\theta}(A(\delta, x, t)), y)) \label{eq:our-update}
\end{equation}
for a sample $x \sim \mathcal{D}$ and transformation $t \sim T$.

\subsection{Experimental Setup}
\label{sec:experimental-setup}

We evaluate on YOLOv3 pretrained for COCO 2014 \cite{coco} ($416 \times 416$ pixels). The implementation of YOLOv3 achieves around $55.4\%$ mAP-50 (mAP at $0.5$ IOU metric) using an object-confidence threshold of $0.001$ for non-max suppression. Because mAP is considerably influenced by this threshold, we also evaluate at the $0.1$ confidence threshold used during validation, as well as the $0.5$ confidence threshold used by default for real-time detection. The implementation achieves $50.3\%$ mAP-50 at the $0.1$ confidence threshold and $40.9\%$ at $0.5$.

We define a ``step" as $1,000$ iterations. The following experiments were run for $300$ steps with an initial learning rate of $0.1$ and momentum of $0.9$, which were chosen heuristically. Learning rate was decayed by $0.95$ every $5$ steps, at which point we also run one validation step for the mAP-50 plots. Because the loss functions are highly non-convex, we take the best of $5$ random restarts to mitigate the effects of local optima. Where applicable, patch transformations involved randomly rotating around the $x,y,z$ axes, randomly scaling and translating, and randomly adjusting the brightness of the patch (converting to HSV and scaling V). 
%Henceforth let mAP and mAP-50 be used interchangeably.

\subsection{Unclipped Attack}
\label{sec:unclipped_untargeted}
For the unclipped attack, our method performs the update in \autoref{eq:our-update}, except without clipping. The purpose is to benchmark against DPatch which uses \autoref{eq:dpatch-method}. For both methods, $t$ scales the patch to a fixed $120 \times 120$ pixels and positions at top-left of the image (as in \cite{dpatch}).

\begin{figure}[t!]
    \centering
    \begin{subfigure}[b]{0.2\textwidth}
        \includegraphics[width=\textwidth]{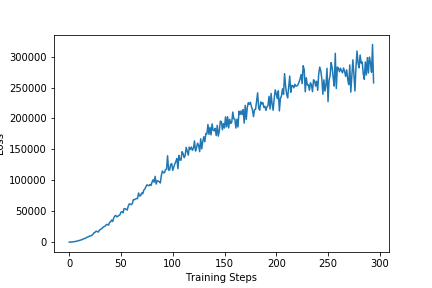}
        \caption{Training Loss (Ours)}
    \end{subfigure}
    ~
    \begin{subfigure}[b]{0.2\textwidth}
        \includegraphics[width=\textwidth]{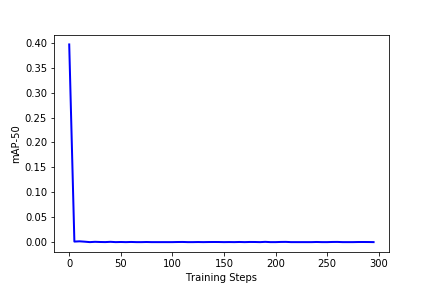}
        \caption{mAP-50 (Ours)}
    \end{subfigure}
    % blank line to force subfigure onto newline
    \begin{subfigure}[b]{0.2\textwidth}
        \includegraphics[width=\textwidth]{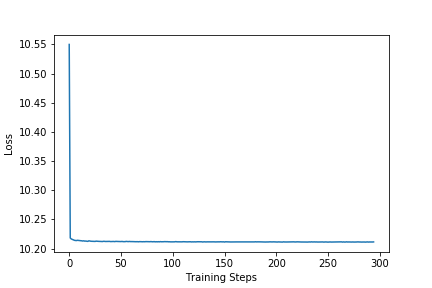}
        \caption{Training Loss (DPatch)}
    \end{subfigure}
    ~
    \begin{subfigure}[b]{0.2\textwidth}
        \includegraphics[width=\textwidth]{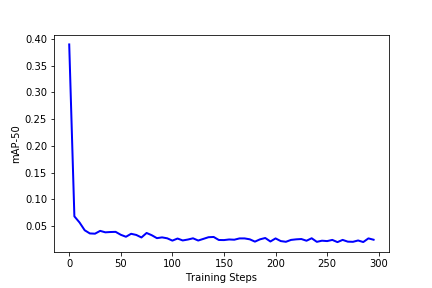}
        \caption{mAP-50 (DPatch)}
    \end{subfigure}
    \caption{Our method vs. DPatch for the unclipped case. }
    \label{fig:unclipped_untargeted_plots}
\end{figure}

\autoref{fig:unclipped_untargeted_plots} shows that our method achieves approximately $0$ mAP after only $5$ steps, whereas DPatch converges to roughly $3$ mAP after $50$ steps. From our experiments, lowering the learning rate or decaying more aggressively does not help to decrease the DPatch mAP, perhaps indicating a limitation in the loss function itself.

\begin{table}[t]\tiny
    \begin{adjustbox}{center}
    \begin{tabular}{ |l*{5}{l}| }
     \hline
     Method & Conf. & mAP (\%) & Smallest AP (\%) & Largest AP (\%) & \\
     \hline
     Baseline & 0.001 & 55.4 & 10.23 (Hair Drier) & 87.61 (Giraffe) & \\
     Baseline & 0.1 & 50.3 & 3.03 (Toaster) & 82.13 (Giraffe) & \\
     Baseline & 0.5 & 40.9 & 0 (Toaster) & 79.53 (Giraffe) & \\
     DPatch & 0.001 & 9.21 & 0.17 (Traffic Light) & 26.07 (Clock) & \\
     DPatch & 0.1 & 7.23 & 0 (Toaster) & 19.83 (Mouse) & \\
     DPatch & 0.5 & 4.88 & 0 (Toaster) & 15.83 (Microwave) & \\
     Ours & 0.001 & \textbf{0.25} & 0 (Aeroplane) & 2.2 (Sports Ball) & \\
     Ours & 0.1 & \textbf{0.1} & 0 (Aeroplane) & 1.2 (Knife) & \\
     Ours & 0.5 & \textbf{0.05} & 0 (Bicycle) & 0.76 (Aeroplane) & \\
     \hline
    \end{tabular}
    \end{adjustbox}
    \caption{Summary for the unclipped case. ``Baseline" is no patch.}
    \label{tab:unclipped_untargeted_table}
\end{table}

\autoref{tab:unclipped_untargeted_table} shows the overall mAP as well as smallest and largest per-class APs for various confidence thresholds. These values were obtained by evaluating on the entire validation set instead of just one ``step". Our DPatch results are mostly consistent with \cite{dpatch} which reports $3.4$ mAP for the untargeted attack on YOLOv2 and Pascal VOC 2007 -- deviations are expected due to differences in implementation, model architecture and dataset. 

\begin{figure}[t!]
    \centering
    \begin{subfigure}[h]{0.15\textwidth}
        \includegraphics[width=\textwidth]{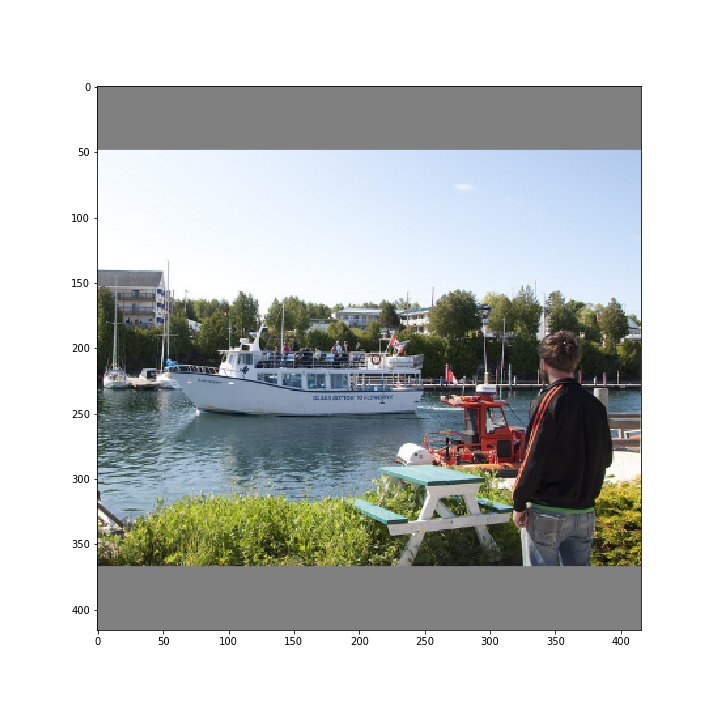}
        % \caption{Original Image}
    \end{subfigure}
    ~
    \begin{subfigure}[h]{0.15\textwidth}
        \includegraphics[width=\textwidth]{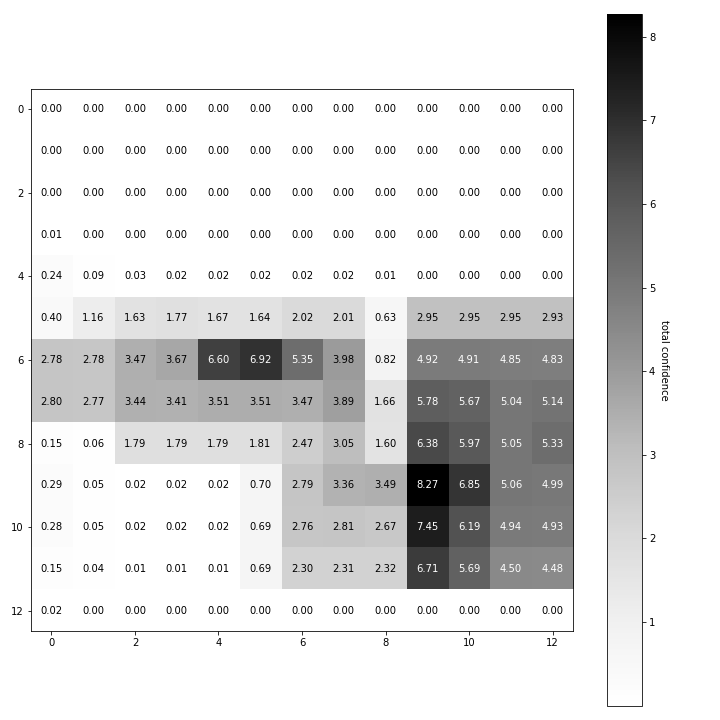}
        % \caption{Original Heatmap}
    \end{subfigure}
    % blank line to force subfigure onto newline
    
    \begin{subfigure}[h]{0.15\textwidth}
        \includegraphics[width=\textwidth]{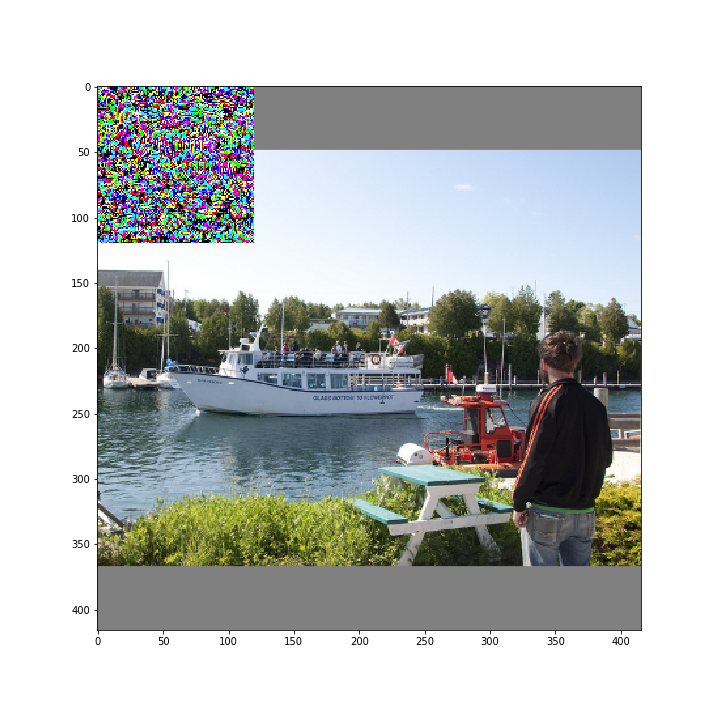}
        % \caption{Attacked Image}
    \end{subfigure}
    ~
    \begin{subfigure}[h]{0.15\textwidth}
        \includegraphics[width=\textwidth]{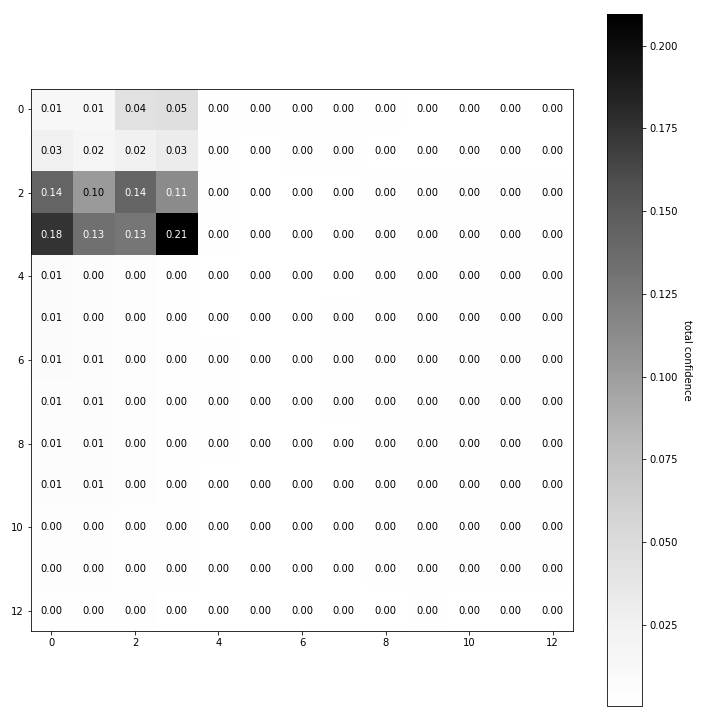}
        % \caption{Attacked Heatmap}
    \end{subfigure}
    \caption{ROI plots for the unclipped case. Top row shows original.}
    \label{fig:unclipped_untargeted_samples}
\end{figure}

To verify that our patch attacks at the bounding box proposal level, we plot the pre-non-max suppression bounding box confidence scores for a random image, shown in \autoref{fig:unclipped_untargeted_samples}.

\subsection{Clipped Attack}
For the clipped attack, our method performs the update from \autoref{eq:our-update}. We compare with DPatch, which uses \autoref{eq:dpatch-method} but modified to clip the patch to $[0,1]$.

\begin{figure}[t!]
    \centering
    \begin{subfigure}[b]{0.2\textwidth}
        \includegraphics[width=\textwidth]{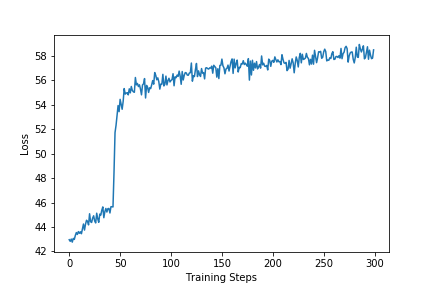}
        \caption{Training Loss (Ours)}
    \end{subfigure}
    ~
    \begin{subfigure}[b]{0.2\textwidth}
        \includegraphics[width=\textwidth]{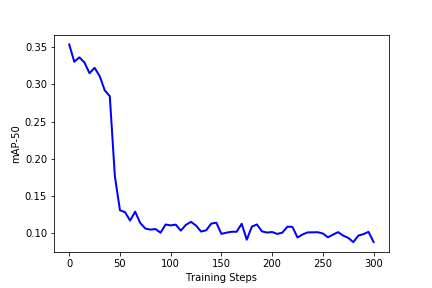}
        \caption{mAP-50 (Ours)}
    \end{subfigure}
    % blank line to force subfigure onto newline
    \begin{subfigure}[b]{0.2\textwidth}
        \includegraphics[width=\textwidth]{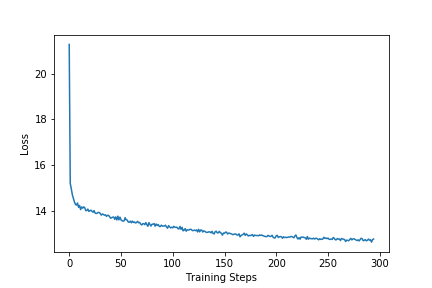}
        \caption{Training Loss (DPatch)}
    \end{subfigure}
    ~
    \begin{subfigure}[b]{0.2\textwidth}
        \includegraphics[width=\textwidth]{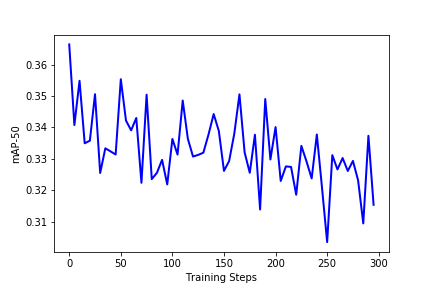}
        \caption{mAP-50 (DPatch)}
    \end{subfigure}
    \caption{Our method vs. DPatch for the clipped case.}
    \label{fig:clipped_untargeted_plots}
\end{figure}

\autoref{fig:clipped_untargeted_plots} shows the loss and mAP plots for a clipped patch with all transforms as described in \autoref{sec:experimental-setup}. Specifically, we randomly rotated $[-5,5]^{\circ}$ for $x,y$ and $[-10,10]^{\circ}$ for $z$; scaled between $80 \times 80$ to $120 \times 120$ pixels; and adjusted brightness by factor from $[0.4, 1.6]$. Translations were sampled post-scaling such that the patch could appear in any location in the image, and scale was adjusted to ensure the patch is not ``cut off" after rotation. 

The DPatch method quickly converges to a patch that is weakly adversarial, whereas our method achieves single-digit mAP values. As in the unclipped case, random restarts and hyperparameter tuning do not appear to help DPatch improve significantly. \autoref{tab:clipped_untargeted_table} shows the AP breakdown as evaluated on the entire validation set, this time applying the patch at random locations in the image.

\begin{figure}[t!]
    \centering
    \begin{subfigure}[h]{0.15\textwidth}
        \includegraphics[width=\textwidth]{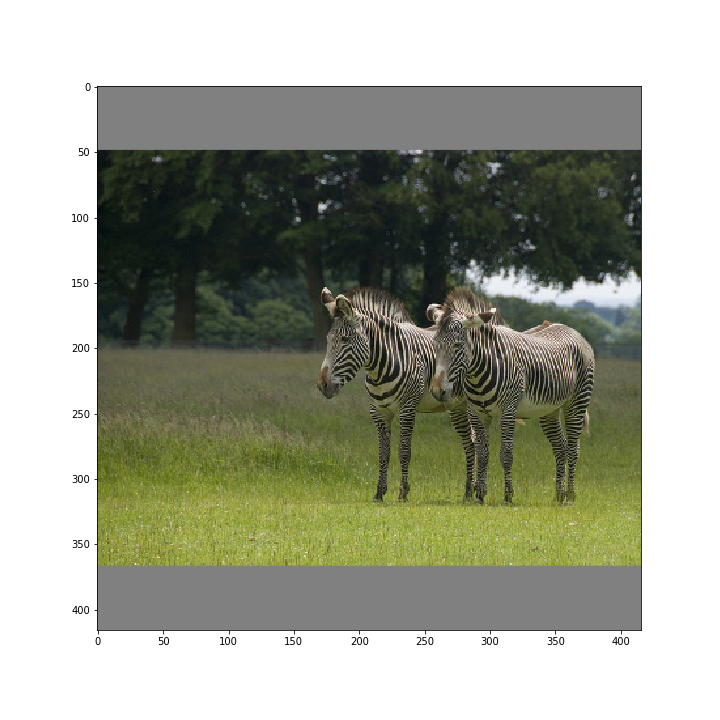}
        % \caption{Original Image}
    \end{subfigure}
    ~
    \begin{subfigure}[h]{0.15\textwidth}
        \includegraphics[width=\textwidth]{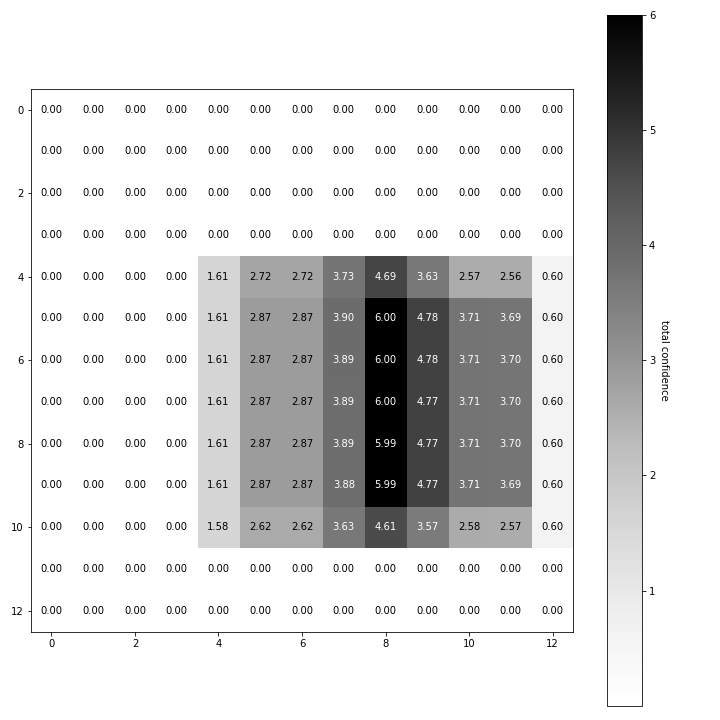}
        % \caption{Original Heatmap}
    \end{subfigure}
    % blank line to force subfigure onto newline
    
    \begin{subfigure}[h]{0.15\textwidth}
        \includegraphics[width=\textwidth]{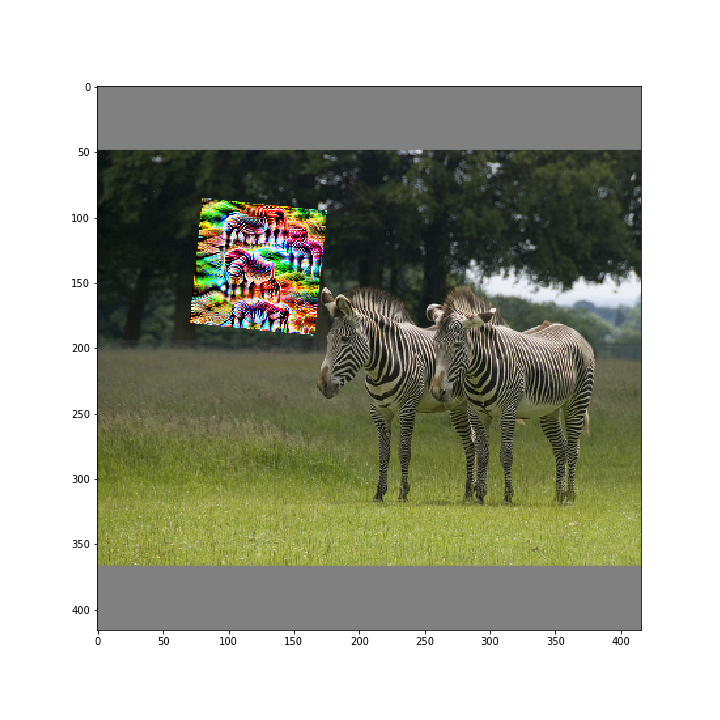}
        % \caption{Attacked Image}
    \end{subfigure}
    ~
    \begin{subfigure}[h]{0.15\textwidth}
        \includegraphics[width=\textwidth]{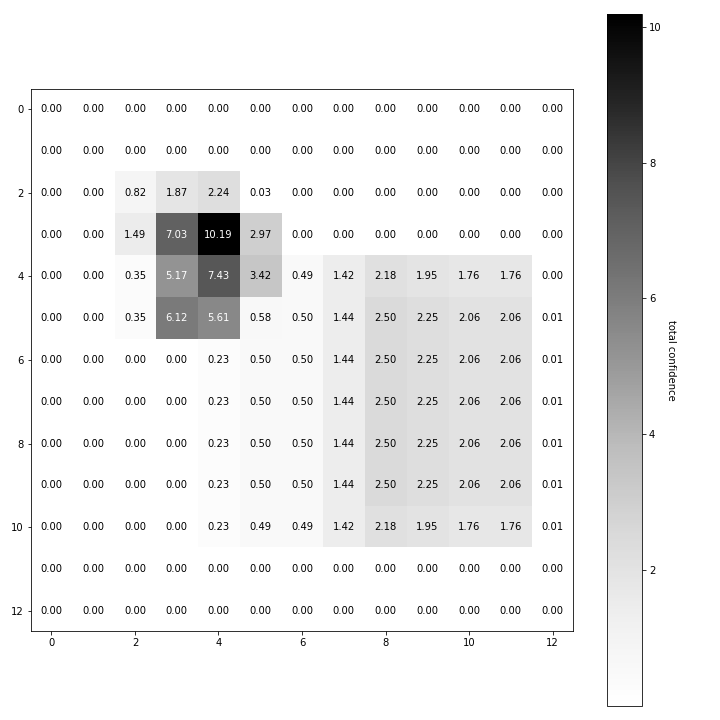}
        % \caption{Attacked Heatmap}
    \end{subfigure}
    \caption{ROI plots for the clipped case. Top row shows original.}
    \label{fig:clipped_untargeted_samples}
\end{figure}

Our patch achieves as low as $7.2$ mAP, almost comparable to an unclipped DPatch. The clipped DPatch is only marginally better than a random image.  Our patch also uniquely captures semantically meaningful patterns (zebra stripes) that are most salient to the detector. Like the unclipped case, \autoref{fig:clipped_untargeted_samples} shows that our patch successfully attracts most of the region proposals. 

\begin{table}[t]\tiny
    \begin{adjustbox}{center}
    \begin{tabular}{ |l*{5}{l}| }
     \hline
     Method & Conf. & mAP & Smallest AP (\%) & Largest AP (\%) & \\
     \hline
     Baseline & 0.001 & 55.4 & 10.23 (Hair Drier) & 87.61 (Giraffe) & \\
     Baseline & 0.1 & 50.3 & 3.03 (Toaster) & 82.13 (Giraffe) & \\
     Baseline & 0.5 & 40.9 & 0 (Toaster) & 79.53 (Giraffe) & \\
     DPatch & 0.001 & 39.6 & 9.09 (Hair Drier) & 69.03 (Train) & \\
     DPatch & 0.1 & 34.7 & 0 (Toaster) & 66.3 (Bus) & \\
     DPatch & 0.5 & 26.8 & 0 (Toaster) & 58.1 (Bus) & \\
     Ours & 0.001 & \textbf{13.8} & 1.27 (Zebra) & 34.69 (Car) & \\
     Ours & 0.1 & \textbf{10.4} & 0 (Toaster) & 30.18 (Car) & \\
     Ours & 0.5 & \textbf{7.2} & 0 (Hot Dog) & 23.06 (Person) & \\
     \hline
    \end{tabular}
    \end{adjustbox}
    \caption{Summary for the clipped case. ``Baseline" is no patch.}
    \label{tab:clipped_untargeted_table}
\end{table}

\begin{figure}[t]
    \centering
    \begin{subfigure}[b]{0.1\textwidth}
        \includegraphics[width=\textwidth]{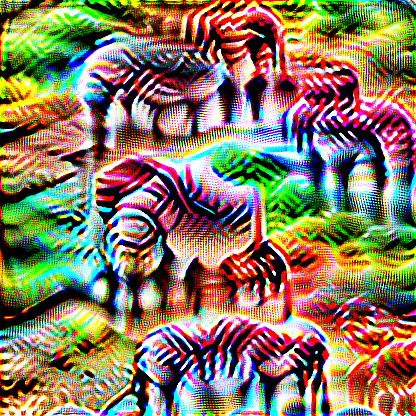}
        \caption{Ours}
    \end{subfigure}
    ~
    \begin{subfigure}[b]{0.1\textwidth}
        \includegraphics[width=\textwidth]{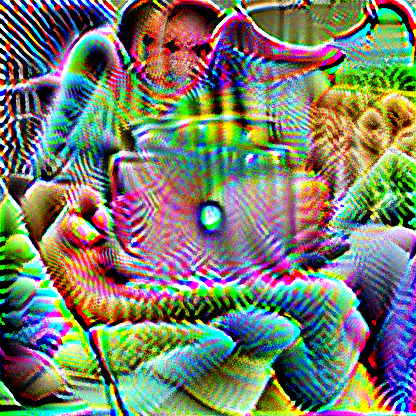}
        \caption{DPatch}
    \end{subfigure}
    \caption{Comparison of patches.}
    \label{fig:clipped_untargeted_patches}
\end{figure}

\subsection{Physical Attack}

\begin{figure}[t!]
    \centering
    \begin{subfigure}[h]{0.13\textwidth}
        \includegraphics[width=\textwidth]{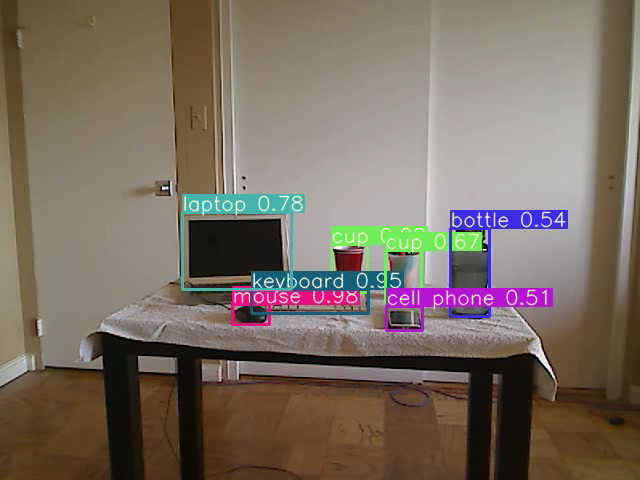}
    \end{subfigure}
    ~
    \begin{subfigure}[h]{0.13\textwidth}
        \includegraphics[width=\textwidth]{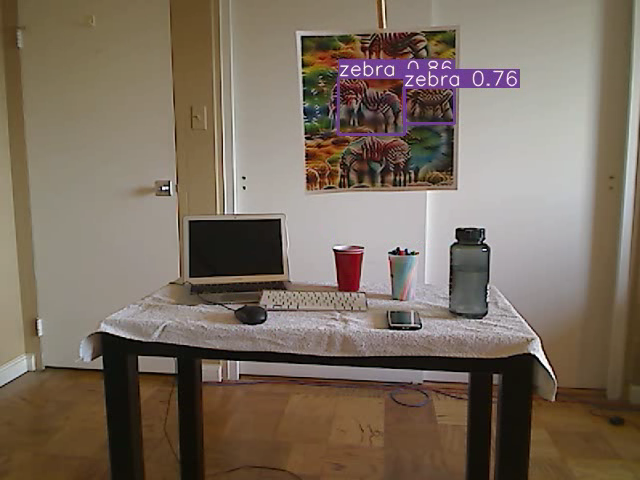}
    \end{subfigure}
    \caption{Physical attack using our patch.}
    \label{fig:physical_demo}
\end{figure}

\begin{figure}[t!]
    \centering
    \begin{subfigure}[h]{0.15\textwidth}
        \includegraphics[width=\textwidth]{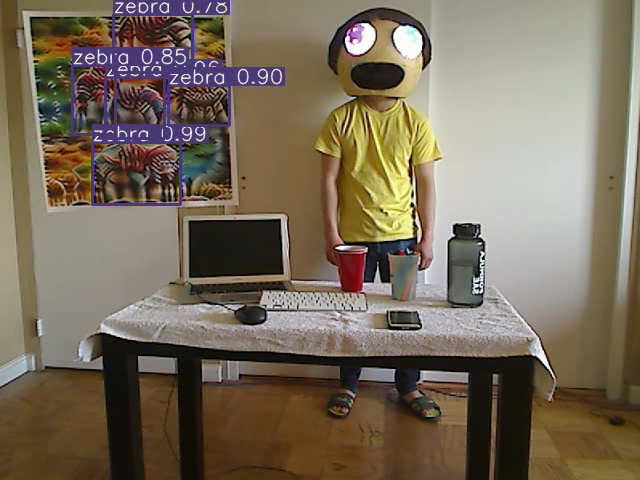}
    \end{subfigure}
    ~
    \begin{subfigure}[h]{0.15\textwidth}
        \includegraphics[width=\textwidth]{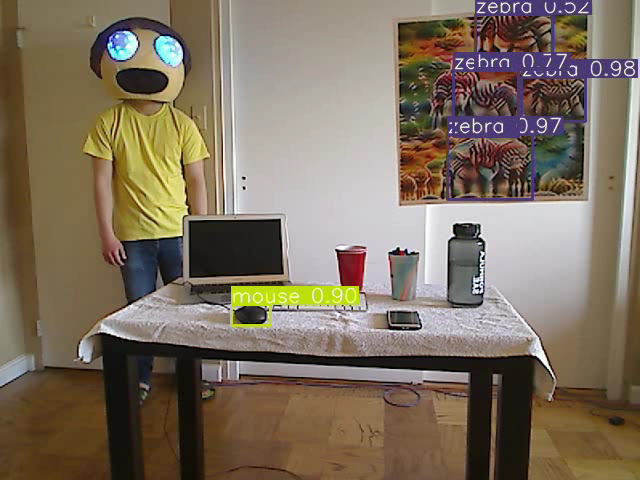}
    \end{subfigure}
    \caption{Location invariance of our patch in physical space.}
    \label{fig:physical_translations}
\end{figure}

\begin{figure}[t!]
    \centering
    \begin{subfigure}[h]{0.13\textwidth}
        \includegraphics[width=\textwidth]{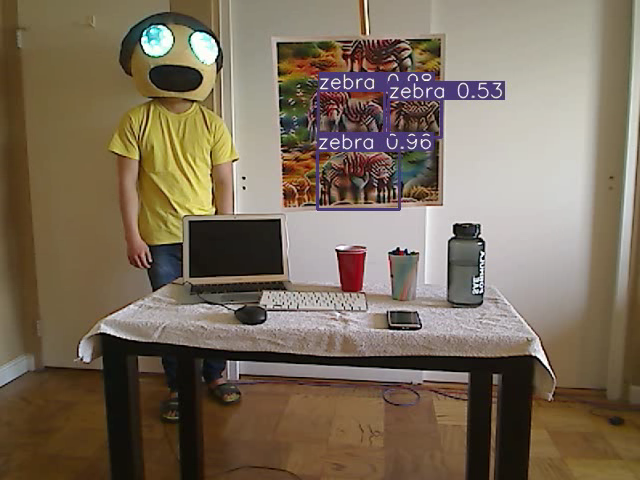}
    \end{subfigure}
    ~
    \begin{subfigure}[h]{0.13\textwidth}
        \includegraphics[width=\textwidth]{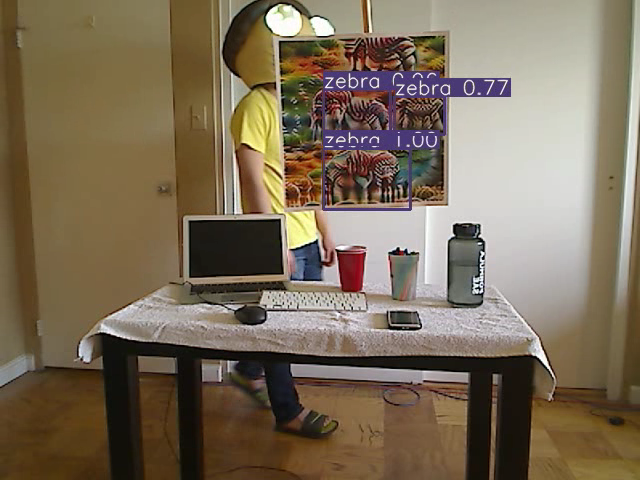}
    \end{subfigure}
    ~
    \begin{subfigure}[h]{0.13\textwidth}
        \includegraphics[width=\textwidth]{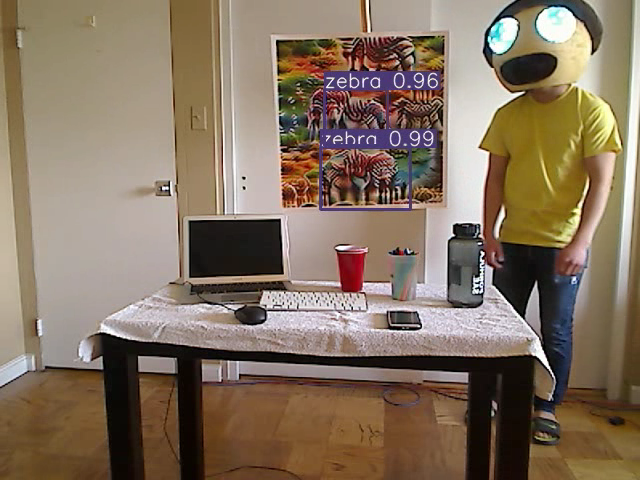}
    \end{subfigure}
    \caption{Moving object suppression in physical space.}
    \label{fig:physical_movement}
\end{figure}

\autoref{fig:physical_demo} shows a printed version of our patch attacking YOLOv3 running real-time with a standard webcam. The patch was printed on regular printer paper and recorded under natural lighting. While the patch is somewhat invariant to location, the patch generally has weaker influence on objects that are farther away, as seen in \autoref{fig:physical_translations} -- when positioned at the sides, the patch needs to be enlarged to successfully disable distant detections, and fails to disable sufficiently confident ones. However, the patch is able to disable detections that are moving, so long as the patch itself is stable, as shown in \autoref{fig:physical_movement}.  This shows that our patch works on a data distribution different from the training distribution, and is generally adversarial over different lighting conditions, positions and orientations. 

\subsection{Discussion}
\label{sec:why-it-works}
We suspect DPatch struggles because it centralizes all ground truth boxes around the patch -- it ultimately resides in a single cell, meaning the loss is dominated by the proposal ``responsible" for that cell. As long as the patch is recognized, the model incurs little penalty for predicting all the other objects, perhaps suffering penalty on the objectness scores but not on bounding boxes or class labels. The loss can there be reduced even if the model behavior does not change much. And in practice, the patch is often detected with high confidence without suppressing other detections. In our method, every grid cell overlapped by a ground truth box contributes to the loss, which increases the most when the model fails to predict any ground truth box.

\section{Conclusion}
We introduce a patch attack causing YOLOv3 to drop from $55.4$ to single digit mAP. We show that this method outperforms the existing DPatch method in the untargeted case, which generally has equivalently significant implications as a targeted attack. Finally, we demonstrate that our attack extends to the physical space by printing our patch and fooling YOLOv3 running real-time via webcam feed, which to our knowledge is the first demonstration of a patch attack on object detectors that successfully suppresses detections without having to overlap the patch and the target objects.

\bibliography{example_paper}

\begin{thebibliography}{15}
\providecommand{\natexlab}[1]{#1}
\providecommand{\url}[1]{\texttt{#1}}
\expandafter\ifx\csname urlstyle\endcsname\relax
  \providecommand{\doi}[1]{doi: #1}\else
  \providecommand{\doi}{doi: \begingroup \urlstyle{rm}\Url}\fi

\bibitem[Athalye et~al.(2018)Athalye, Engstrom, Ilyas, and
  Kwok]{athalye2018synthesizing}
Athalye, A., Engstrom, L., Ilyas, A., and Kwok, K.
\newblock Synthesizing robust adversarial examples.
\newblock In \emph{International Conference on Machine Learning}, pp.\
  284--293, 2018.

\bibitem[Bose \& Aarabi(2018)Bose and Aarabi]{adversarial-face}
Bose, A.~J. and Aarabi, P.
\newblock Adversarial attacks on face detectors using neural net based
  constrained optimization.
\newblock \emph{CoRR}, abs/1805.12302, 2018.
\newblock URL \url{http://arxiv.org/abs/1805.12302}.

\bibitem[Brown et~al.(2017)Brown, Man{\'{e}}, Roy, Abadi, and
  Gilmer]{adversarial-patch}
Brown, T.~B., Man{\'{e}}, D., Roy, A., Abadi, M., and Gilmer, J.
\newblock Adversarial patch.
\newblock \emph{CoRR}, abs/1712.09665, 2017.
\newblock URL \url{http://arxiv.org/abs/1712.09665}.

\bibitem[Chen et~al.(2018)Chen, Cornelius, Martin, and Chau]{shapeshifter}
Chen, S., Cornelius, C., Martin, J., and Chau, D.~H.
\newblock Robust physical adversarial attack on faster {R-CNN} object detector.
\newblock \emph{CoRR}, abs/1804.05810, 2018.
\newblock URL \url{http://arxiv.org/abs/1804.05810}.

\bibitem[Eykholt et~al.(2018)Eykholt, Evtimov, Fernandes, Li, Rahmati, Xiao,
  Prakash, Kohno, and Song]{eykholt2018robust}
Eykholt, K., Evtimov, I., Fernandes, E., Li, B., Rahmati, A., Xiao, C.,
  Prakash, A., Kohno, T., and Song, D.~X.
\newblock Robust physical-world attacks on deep learning visual classification.
\newblock \emph{2018 IEEE/CVF Conference on Computer Vision and Pattern
  Recognition}, pp.\  1625--1634, 2018.

\bibitem[Kurakin et~al.(2017)Kurakin, Goodfellow, and
  Bengio]{kurakin2016adversarial}
Kurakin, A., Goodfellow, I., and Bengio, S.
\newblock Adversarial machine learning at scale.
\newblock In \emph{International Conference on Learning Representations}, 2017.

\bibitem[Kurakin et~al.(2018)Kurakin, Goodfellow, and
  Bengio]{kurakin2018adversarial}
Kurakin, A., Goodfellow, I.~J., and Bengio, S.
\newblock Adversarial examples in the physical world.
\newblock In \emph{Artificial Intelligence Safety and Security}, pp.\  99--112.
  Chapman and Hall/CRC, 2018.

\bibitem[Lin et~al.(2014)Lin, Maire, Belongie, Bourdev, Girshick, Hays, Perona,
  Ramanan, Doll{\'{a}}r, and Zitnick]{coco}
Lin, T., Maire, M., Belongie, S.~J., Bourdev, L.~D., Girshick, R.~B., Hays, J.,
  Perona, P., Ramanan, D., Doll{\'{a}}r, P., and Zitnick, C.~L.
\newblock Microsoft {COCO:} common objects in context.
\newblock \emph{CoRR}, abs/1405.0312, 2014.
\newblock URL \url{http://arxiv.org/abs/1405.0312}.

\bibitem[Liu et~al.(2018)Liu, Yang, Song, Li, and Chen]{dpatch}
Liu, X., Yang, H., Song, L., Li, H., and Chen, Y.
\newblock Dpatch: Attacking object detectors with adversarial patches.
\newblock \emph{CoRR}, abs/1806.02299, 2018.
\newblock URL \url{http://arxiv.org/abs/1806.02299}.

\bibitem[Madry et~al.(2017)Madry, Makelov, Schmidt, Tsipras, and
  Vladu]{madry2017towards}
Madry, A., Makelov, A., Schmidt, L., Tsipras, D., and Vladu, A.
\newblock Towards deep learning models resistant to adversarial attacks.
\newblock In \emph{International Conference on Learning Representations}, 2017.

\bibitem[Redmon \& Farhadi(2018)Redmon and Farhadi]{yolov3}
Redmon, J. and Farhadi, A.
\newblock Yolov3: An incremental improvement.
\newblock \emph{CoRR}, abs/1804.02767, 2018.
\newblock URL \url{http://arxiv.org/abs/1804.02767}.

\bibitem[Sharif et~al.(2016)Sharif, Bhagavatula, Bauer, and
  Reiter]{sharif2016accessorize}
Sharif, M., Bhagavatula, S., Bauer, L., and Reiter, M.~K.
\newblock Accessorize to a crime: Real and stealthy attacks on state-of-the-art
  face recognition.
\newblock In \emph{Proceedings of the 2016 ACM SIGSAC Conference on Computer
  and Communications Security}, pp.\  1528--1540. ACM, 2016.

\bibitem[Sharif et~al.(2018)Sharif, Bhagavatula, Bauer, and
  Reiter]{adversarial-glasses}
Sharif, M., Bhagavatula, S., Bauer, L., and Reiter, M.~K.
\newblock Adversarial generative nets: Neural network attacks on
  state-of-the-art face recognition.
\newblock \emph{CoRR}, abs/1801.00349, 2018.
\newblock URL \url{http://arxiv.org/abs/1801.00349}.

\bibitem[Thys et~al.(2019)Thys, Ranst, and Goedem{\'{e}}]{surveillance}
Thys, S., Ranst, W.~V., and Goedem{\'{e}}, T.
\newblock Fooling automated surveillance cameras: adversarial patches to attack
  person detection.
\newblock \emph{CoRR}, abs/1904.08653, 2019.
\newblock URL \url{http://arxiv.org/abs/1904.08653}.

\bibitem[Xie et~al.(2017)Xie, Wang, Zhang, Zhou, Xie, and
  Yuille]{xie2017adversarial}
Xie, C., Wang, J., Zhang, Z., Zhou, Y., Xie, L., and Yuille, A.
\newblock Adversarial examples for semantic segmentation and object detection.
\newblock In \emph{Proceedings of the IEEE International Conference on Computer
  Vision}, pp.\  1369--1378, 2017.

\end{thebibliography}
\bibliographystyle{icml2019}

\end{document}